\documentclass[twocolumn,10pt]{IEEEtran}
\usepackage{bm,cite,float,amsmath,amssymb,amsthm}
\usepackage{multirow}
\usepackage{algorithm}
\usepackage[noend]{algpseudocode}
\usepackage{indentfirst}
\usepackage{xcolor}
\usepackage{makecell}
\usepackage{color}

\usepackage{lipsum}
\usepackage{mathtools}
\usepackage{cuted}

\makeatletter
\def\BState{\State\hskip-\ALG@thistlm}
\makeatother

\usepackage{amssymb}
\usepackage{amsmath}
\usepackage{graphicx}
\usepackage{cite}

\usepackage{balance}
\usepackage[utf8]{inputenc}

\IEEEoverridecommandlockouts

\usepackage{graphicx,epstopdf}
\usepackage{epsfig}	
\usepackage{amsfonts,balance}
\usepackage{bbm}
\floatname{algorithm}{Algorithm}

\usepackage{lipsum}

\newtheorem{theorem}{Theorem}

\newtheorem{corollary}{Corollary}

\makeatletter
\def\ScaleIfNeeded{%
	\ifdim\Gin@nat@width>\linewidth \linewidth \else \Gin@nat@width
	\fi } \makeatother

\begin{document}
	
	%
	\title{Machine Learning for Massive \\Industrial Internet of Things }
	\author{Hui~Zhou,
		Changyang~She,
		Yansha~Deng,
		Mischa Dohler,
		and Arumugam Nallanathan
		\thanks{Hui Zhou, Yansha Deng, and Mischa Dohler are with the Department of Engineering, King’s College London (e-mail: \{hui.zhou, yansha.deng, mischa.dohler\}@kcl.ac.uk) (Corresponding author: Yansha Deng).}
		\thanks{Changyang She is with the School of Electrical and Information Engineering, University of Sydney (e-mail: changyang.she@sydney.edu.au).}
		\thanks{Arumugam Nallanathan is with the School of Electronic Engineering and Computer Science, Queen Mary University of London (e-mail: a.nallanathan@qmul.ac.uk).}
	}
	\maketitle

	\begin{abstract}
		Industrial Internet of Things (IIoT) revolutionizes the future manufacturing facilities by integrating the Internet of
		Things technologies into industrial settings. With the deployment of massive IIoT devices, it is difficult for the wireless network to support the ubiquitous connections with diverse quality-of-service (QoS) requirements. Although machine learning is regarded as a powerful data-driven tool to optimize wireless network, how to apply machine learning to deal with the massive IIoT problems with unique characteristics remains unsolved. In this paper, we first summarize the QoS requirements of the typical massive non-critical and critical IIoT use cases. We then identify unique characteristics in the massive IIoT scenario, and the corresponding machine learning solutions with its limitations and potential research directions. We further present the existing machine learning solutions for individual layer and cross-layer problems in massive IIoT.  Last but not the least, we present a case study of massive access problem based on deep neural network and deep reinforcement learning techniques, respectively,  to validate the effectiveness of machine learning in massive IIoT scenario.  
	\end{abstract}
	
	

	%
	\maketitle

	\section{Introduction}
	
	Industrial Internet of Things (IIoT), also referred to as Industrial 4.0, integrates IoT technologies into the industrial field, which will revolutionize manufacturing, data analysis, and logistic process in smart factories. As IIoT rapidly evolves, dramatically increasing number of devices impose high demands on the existing cellular network, with the expectation to support the ubiquitous connections from both critical and non-critical IoT devices. To reach this expectation, massive Machine Type Communications service has been standardized in the Fifth Generation (5G) New Radio (NR) for up to $10^6/\mathrm{km^2}$ massive non-critical IoT devices \cite{jiang2020traffic}, and massive Ultra Reliable Low Latency Communications are envisioned to be one of the services in Six Generation (6G) to support massive critical IoT devices with 1ms user plane latency target.

	Optimizing the operation of the cellular network to meet the diverse quality-of-service (QoS) requirements is challenging in a large-scale IIoT environment, where a collection of heterogeneous devices are geographically distributed. Despite the remarkable success of the traditional optimization methods (e.g., convex optimization), it turns out that most wireless optimization problems in the IIoT environment are non-convex, which end up with locally optimal solutions. Another limitation of traditional optimization methods lies in its requirement of exact models, which are difficult to obtain in the dynamic IIoT environment with diverse QoS requirements. 
	
	Although machine learning is regarded as a powerful data-driven method to enable intelligent decision making by utilizing the high volume of data from heterogeneous IIoT devices. The unique characteristics in massive IIoT limit the application of machine learning algorithms, which cannot guarantee the optimal decision under stringent constraints or even convergence. Several existing works have focused on specific machine learning solutions for IoT problems \cite{jiang2020traffic, song2020artificial, khan2020federated}. In \cite{jiang2020traffic}, the access control algorithms have been proposed to optimize the massive access problem based on the deep reinforcement learning (DRL). In \cite{song2020artificial}, centralized and decentralized IoT frameworks with DRL solutions have been exploited. In \cite{khan2020federated}, an federated learning (FL) incentive mechanism has been proposed for resource optimization. Yet, a comprehensive study on the unique characteristics in the massive IIoT scenario, and potential machine learning solutions has never been carried out.

	\begin{figure*}[!tb]
		\centerline{\includegraphics[scale=0.55]{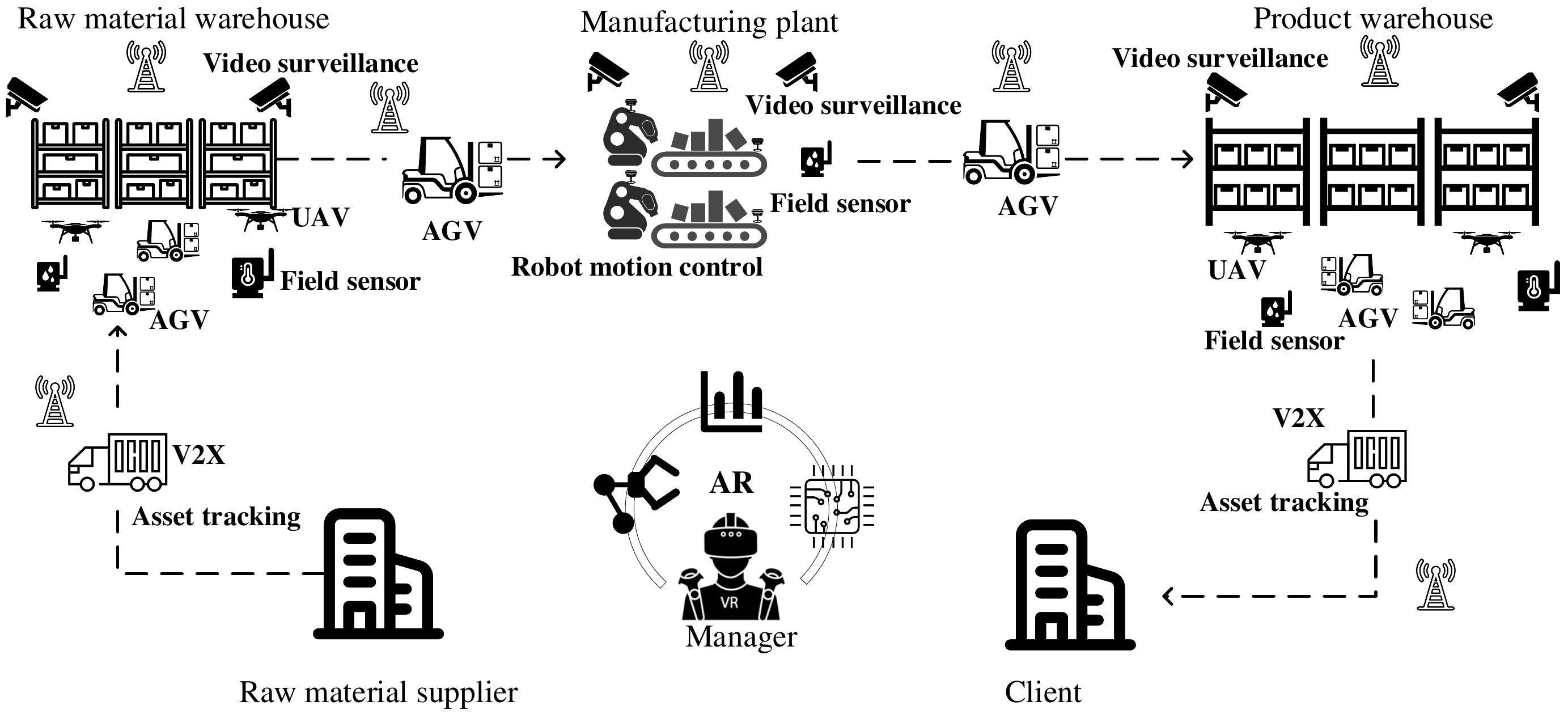}}
		\caption{A typical 5G and Beyond factory in the future.}
		\label{fig:typical factory in the future}
	\end{figure*}	
	\begin{table*}[tb]
		\caption{Typical  QoS  Requirements  of  Different Use Cases in Smart Factory}
		\begin{center}
			\begin{tabular}{|l|c|c|c|c|c|}
				\hline
				Use case & Reliability & Latency & Data Rate&Connection Density& Category\\
				\hline			
				Field sensors& 99.9\% &10sec &10kbps & 1,000,000/$\mathrm{km}^2$& \multirow{3}{*}{Massive non-critical} \\
				\cline{1-5}
				Asset tracking& 99.9\% &10sec &0.1-20kbps & 1,000,000/$ \mathrm{km}^2 $&  \\
				\cline{1-5}
				Video surveillance&99.9\% &100ms &1-10Mbps &1000/$\mathrm{km}^2$&  \\
				\hline
				Vehicle-to-everything communications &99.999\% &10ms & 20Mbps&-&\\
				\cline{1-5}
				Automated guided vehicles &99.9999\% & 1-50ms& 1-10Mbps&-&  \\
				\cline{1-5}
				Unmanned aerial vehicles& 99.9999\%& 1-50ms&1-10Mbps &-&Massive critical \\
				\cline{1-5}
				Augmented reality & 99.9999\%&1-5ms &5-25Mbps &-& \\
				\cline{1-5}		
				Robot motion control& 99.9999\%& 1ms&1-10Mbps &-& \\
				\hline
			\end{tabular}
			\label{QoS requirements}
		\end{center}
	\end{table*}
	
	The main contributions of this paper are: 1) we first summarize the diverse QoS requirements of massive IIoT use cases in Section II; 2) we then provide a concrete vision of the unique characteristics in massive IIoT scenario, and investigate machine learning solutions along with its limitations and potential research directions in Section~III; 3) we present how different machine learning algorithms can be applied in individual layer and cross-layer design to guarantee the QoS requirements of massive IIoT in Section IV; 4) to demonstrate the effectiveness of machine learning in massive IIoT scenario, we present a case study of massive access optimization problem, specifically the unsupervised learning with different deep neural network (DNN) structures and DRL, respectively, and analyze the results in Section V. Finally, we conclude the paper in Section VI.

	\section{Massive Industrial IoT Application Scenarios}
	
	As illustrated in Fig.~\ref{fig:typical factory in the future}, 5G and Beyond is expected to support the smart factory to guarantee the operation efficiency. In this section, we summarize the QoS requirements of massive non-critical and massive critical IIoT use cases as shown in Table~\ref{QoS requirements}\footnote{In Industrial 4.0, the density of devices such as AGVs in indoor applications will be extremely high (i.e., $10^6/\mathrm{km^2}$), and the supply chains could be global.}, and point out their specific challenges.

	\subsection{Massive Non-critical IoT}
	Massive low/medium-end IoT devices are deployed in the factory to support the manufacturing by collecting temperature, position, and etc. Although they have much lower latency and reliability requirements, the density of devices can be up to $\mathrm{10^6/km^2}$.
	\subsubsection{Field Sensors}
	To monitor and support manufacturing, increasing number of sensors and actuators will be deployed in the smart factories to guarantee safety. Although the data rate of a single field sensor is quite low (e.g.,10kbps), the collision from the simultaneous massive access may lead to severe access delays and packet losses.

	\subsubsection{Asset Tracking}
	Accurate locations and inventory status of the assets are critical for operational efficiency and safety. According to requirements in the 5G evolution, we should provide centimeter-level positioning for a large number of devices in IIoT.

	\subsubsection{Video Surveillance}
	Video surveillance is widely adopted in the factory to guarantee warehouse security, and automation process safety, which generates Gbps-level uplink traffic from hundreds of connected cameras. Thus, the uplink enhancements and techniques are critical for the IIoT use case.

	\subsection{Massive Critical IoT}
	In industrial 4.0, supporting ultra-reliable and low-latency communications, e.g., $\mathrm{10^{-5}\sim 10^{-7}}$ packet loss probability and $\mathrm{1ms}$ end-to-end delay, is of significant importance for massive critical IoT applications, such as Automated Guided Vehicles (AGV), Unmanned Aerial Vehicle (UAV), etc.

	\subsubsection{Vehicles} As the typical mission-critical applications, Vehicle-to-everything (V2X) and AGV communications improve the asset transportation efficiency in outdoor and indoor environments, respectively. UAVs are important in applications, such as aerial monitoring, package delivery, etc. The decision making based on the observations of the vehicles, road-sides units, and BSs belongs to sequential decision-making problems, which have no closed-form solutions in general.
	
	\subsubsection{Virtual Environment of Thing} The goal of integrating Augmented Reality (AR) and IoT is to provide real-time control of real-world IoT devices and virtual-world objects in a virtual environment. Apart from the latency and reliability requirements, AR applications need high computational capability and sufficient bandwidth to provide an immersed experience to users.

	\subsubsection{Robot Motion Control}
	Due to the lack of flexibility in robots, human knowledge is indispensable in complex manufacturing processes, where robots respond to human instructions. Due to the fast-changing environment (e.g., delayed positions), the stringent latency and reliability requirements should be satisfied to guarantee the effectiveness of real-time decision making.

	\section{Unique characteristics and Solutions \\in massive IIoT}
	\label{IIoT_challenegs}
	To meet the requirement of massive access, especially under stringent constraints, machine learning is regarded as a powerful tool to optimize the cellular-enabled massive IIoT, which can learn efficient representations of the data. However, the unique characteristics in the massive IIoT scenario limit the application of machine learning algorithms, which include
	\begin{itemize}
		\item temporal data correlation
		\item scalability and high-dimensional data
		\item dynamic networks
		\item limited data
		\item high communication overheads
		\item sequential decision-making problems.
	\end{itemize}
	To deal with these characteristics in massive IIoT, we investigate the potential machine learning solutions along with its limitations, and future research directions in the following, which is also summarized in Table~\ref{DNN solution}\footnote{As a basic universal neural network structure, the Fully-connected Neural Network (FNN), whose neurons in adjacent layers are fully connected, cannot adapt to the unique characteristics in massive IIoT.} and Table~\ref{DNN_limitation}.
	
	\subsection{Temporal Data Correlation}
	IIoT devices produce massive sequential data, which violates the Independent and identically distributed (IID) assumption in traditional machine learning. Taking the data traffic flow as an example, the number of packets transmitted at each time slot is determined by both the number of newly arrived packets and the number of devices with failure transmissions in the previous time slots.

	{\bf{Recurrent Neural Network:}} To capture the temporal data dependencies, recurrent neural network (RNN) with recurrent connections between hidden units can be a promising solution. However, the implementation of RNN in IIoT faces several challenges: 1)  we need to find an optimal memory size of RNN to achieve a good trade-off between the learning performance and computational overheads, however the data traffic correlation time is dynamic in IIoT use cases; 2) for IIoT applications, such as localization, both the information from the past and the future are useful. To extract the two kinds of information, the bidirectional RNN with both forward and backward directions can be applied. It is noted that BRNN needs to wait for the future information to execute the backward pass, which may violate the stringent latency requirement in massive-critical IoT; and 3) the number of parameters in RNN grows rapidly with the number of devices in massive IIoT scenario. As a result, scalability remains an open issue when using RNN in massive IIoT.

	\subsection{Scalability and High Dimensional Data}
	In factory automation scenario, the state and action spaces of the network grow with the number of IIoT devices. Existing machine learning techniques work well in small- and medium-scale problems, but can hardly solve large-scale problems in a reasonable time due to the curse of dimensionality, where feature extraction is needed to reduce the dimensionality and obtain the valuable information from the raw data.

	{\bf{Convolutional Neural Network:}} To solve large-scale problems and extract features from high-dimensional data, we can use convolutional neural network (CNN). With sparse interactions among different layers and commonly used parameters in the filters, the number of parameters in CNN does not change with the scale of the problem. The challenges of applying CNN in IIoT use cases lie in the following two aspects: 1) although CNN has achieved success in certain applications (e.g., the channel estimation), the convolution operation is time-consuming and can not meet the stringent latency requirement in massive-critical IoT; and 2) the CNN can only deal with data in Euclidean space (i.e., 1D sequence or 2D grid), which cannot capture the topology of the wireless network exactly.
	
	{\bf{Graph Neural Network:}} As a generalization of CNN, graph neural network (GNN) is proposed to solve the large-scale and high-dimensional problems with Non-Euclid data structure via aggregating the neighborhood information of each node. As such, it can exploit the underlying topology of wireless networks\cite{rusek2020routenet}. However, there are several challenges when applying GNN in IIoT scenario: 1) to apply GNN at the central server in network optimization, we need to collect the information of all the nodes and edges. This may result in considerable communication overheads and latency in massive IIoT systems; and 2) due to the over-smoothing problem, the performance of GNN gradually decreases with the increasing number of layers. This is because repeated graph convolutions eventually make node embeddings indistinguishable. To enable multi-hop information aggregation in massive IIoT, a deep GNN needs to be exploited.

	\begin{table*}[tb]
		\caption{{Unique characteristics in massive IIoT and machine learning solutions}}
		\begin{center}
			\begin{tabular}{|l|c|c|c|c|c|c|c|c|}
				\hline
				& FNN & RNN & CNN & GNN & GAN & FL & Few-shot learning& DRL\\
				\hline			
				Scalability& &&$\surd$ &$\surd$ &  &&& \\
				\hline
				High dimensional data&& & $\surd$&$\surd$ &  &&& \\
				\hline
				Temporal data correlation&&$\surd$& & & &&& \\
				\hline
				Sequential decision-making problem && & & & & & &$\surd$\\
				\hline
				Dynamic networks && & & $\surd$& &&$\surd$& \\
				\hline
				Limited data&& & & &$\surd$ &&$\surd$&\\
				\hline
				High communication overheads &&& & & &$\surd$&& \\
				\hline		
			\end{tabular}
			\label{DNN solution}
		\end{center}
	\end{table*}

	\subsection{Dynamic Networks}\label{DynNet}
	Wireless networks are highly dynamic such as number of devices, and types of services. A well-trained learning algorithm can hardly guarantee the QoS requirements in dynamic networks. How to fine-tune deep learning algorithms for a new task or in a new environment with few training samples remains a challenging task.
	
	{\bf{Graph Neural Network:}} In dynamic wireless networks, the topology, the number of devices, and the traffic loads change over time. Since offline trained GNN can be transferred to networks unseen in the training, it can be a promising tool for dynamic networks \cite{rusek2020routenet}.
	
	{\bf{Few-shot Learning:}} The other potential solution is few-shot learning, whose motivation is to train a DNN with a small number of samples, and thus the DNN can adapt to dynamic networks without extensive retraining \cite{sun2019meta}. Transfer learning is a widely used few-shot learning method that fine-tunes the pre-trained DNN without changing the hyper-parameters of the DNN. However, the performances of deep learning algorithms are sensitive to hyper-parameters. To handle this issue, meta-learning has been applied in the existing researches to optimize hyper-parameters and has been demonstrated to achieve faster adaptations than transfer learning. The bottlenecks of implementing meta-learning in mission-critical IoT lies in the following two aspects: 1) to enable fast online adaptation, we need vast computational resources to train meta-learning offline, e.g., using hundreds of GPUs to pre-train DNNs in a few days; and 2) although meta-learning can achieve a classification accuracy of more than $70$\% in image classification \cite{sun2019meta}, the error probability is still too high to meet the requirement of mission-critical IoT.

	\subsection{Limited Data}
	To collect enough real-world data samples for machine learning, it takes a very long time. For example, the packet rate of a device is less than $10^3$~packets/s in most use cases. When the reliability requirement is $99.9999$\%, it takes more than $10^3$ seconds to obtain a valid label. As a result, more than $10^7$ seconds are needed to collect $10^4$ labeled training samples from the critical IoT device.
	
	{\bf{Few-shot learning:}} As mentioned in Section \ref{DynNet}, few-shot learning has the potential to enable fast adaptation of deep learning in dynamic networks. The basic idea is to pre-train a general DNN offline and fine-tune it to a specific task with limited real-world data samples. Thus, it is a promising technique to handle different tasks in massive-critical IoT (e.g., UAV) with limited training samples. 
	
	{\bf{Generative Adversarial Network:}} The other promising data augmented method is to use Generative Adversarial Network (GAN) to generate synthetic data samples based on limited real data samples \cite{yang2019generative}. With enough synthetic samples of channel response, packet size, and inter-arrival time between packets, we can evaluate the reliability and pre-train machine learning algorithms. To implement GAN in massive IIoT networks, there are two open challenges: 1) the distribution of synthetic data samples is not exactly the same as that of the real data samples \cite{yang2019generative}. Novel techniques for reducing the gap between the distributions of synthetic and real data are of great interest since they can improve the performance of GAN directly; and 2) for mission-critical IoT, the QoS is very sensitive to the distributions of packet sizes, inter-arrival time between packets, and channel fading. The approximation errors of synthetic data samples may lead to estimation errors of the QoS. Yet, how to quantify their relationship has not been investigated in the existing literature and deserves further study.
	
	\subsection{High Communication Overheads}
	Collecting data samples at the central servers and updating learning parameters at the IIoT devices or base stations will bring high communication overheads in wireless networks. By deploying computing and storage resources at the network edge, it is possible to train and execute distributed learning algorithms with low communication overheads at the cost of longer convergence time. How to improve the tradeoff between communication overheads and the convergence time of distributed learning algorithms is of crucial importance.
	
	{\bf{Federated Learning:}} FL framework has been considered as a promising approach to reduce communication overheads and preserve data privacy. In each FL communication round, devices upload gradients obtained from local data samples to a parameter server. Then, the parameter server aggregates the gradients and updates the DNN by using a gradient descent algorithm. Finally, the DNN is broadcast to all the devices.  However, there are several challenges in deploying the FL system in massive IIoT: 1) the wireless links are unreliable, and the computation capability of each IIoT device is limited. To implement FL over wireless networks with limited local computation capability, over-the-air aggregation and split learning can be potential solutions \cite{kairouz2019advances}. Nevertheless, how to reduce the convergence time and improve the performance of FL by optimizing user scheduling, quantization accuracy, bandwidth allocation, and power control policies remains an open problem; and 2) considering that different devices are operating over different environments and/or in different time windows, the observed data samples can be non-IID data, which makes it more difficult to converge.

	\begin{table}[!t]
		\caption{{Research directions of machine learning in massive IIoT}}
		\begin{center}
			\begin{tabular}{|c|c|}
				\hline
				\textbf{Algorithm} & \textbf{Research direction} \\
				\hline
				GNN  & \makecell[l]{$\bullet$ Performance gain via utilizing network topology\\ $\bullet$ Transference in dynamic networks \\$\bullet$ Cooperation in multi-agent settings}\\
				\hline
				FL & \makecell[l]{$\bullet$ Scheduling policy\\$\bullet$ Cooperation in multi-agent settings}\\
				\hline
				GAN & \makecell[l]{$\bullet$ Quantify the QoS error from synthetic data\\$\bullet$ Tail probability of delay with DRL}\\
				\hline
				\makecell[c]{Few-shot\\ Learning} & \makecell[l]{$\bullet$ Transference in dynamic networks\\ $\bullet$ Non-stationary environment in multi-agent settings}\\
				\hline		
				DRL & \makecell[l]{$\bullet$ Constrained DRL to fulfill QoS requirement\\ $\bullet$ Bellman equation for fast-changing environment\\$\bullet$ Multi-agent DRL}\\
				\hline
			\end{tabular}
			\label{DNN_limitation}
		\end{center}
	\end{table}
	\subsection{Sequential Decision-making Problems}
	Finding the optimal solutions to sequential decision-making problems is very challenging, since the states and actions in different time steps are correlated with each other. To characterize the correlation, one approach is to model the transitions among states by a Markov decision process (MDP). In massive IIoT, each device does not have a full observation of the environment due to the hardware limitation. In this case, MDP becomes Partially Observable Markov Decision Process (POMDP). Solving POMDP problems is more challenging than MDP problems, since the device does not know whether an action is optimal or not from the global view. Finally, each device in massive critical IoT use cases has a stringent QoS requirement, which brings unprecedented challenges, even in MDP problems.

	{\bf{Deep Reinforcement Learning:}}	
	Reinforcement learning (RL) is a machine learning technique to solve MDP problems, where the agent chooses the optimal action to maximize the long-term reward through interacting with the environment. However, traditional RL methods do not scale well in massive IIoT with high dimensional state-action spaces. Therefore, DRL, the combination of RL with the neural network, is regarded as a promising tool in complex POMDP problems. Deep Q-Network (DQN) is a classic DRL algorithm, which estimates the Q value via a neural network. Based on double DQN, Deep Deterministic Policy Gradients (DDPG) is further proposed, which integrates DQN with the actor-critic structure to deal with the continuous action space. The implementation of DRL in a realistic IIoT environment still faces several challenges: 
	\begin{itemize}
		\item {\bf{Multi-objective optimization}}
		In IIoT applications, multiple performance metrics need to be jointly optimized, and hence the problem is generally multi-objective optimization. Most of the existing studies maximize/minimize a weighted sum of different performance metrics, where the weighting coefficients are determined manually. To deal with it, one possible solution is the constrained DRL, where the latency and reliability requirements are formulated as constraints. Then, the optimization variables and the weighting coefficients can be optimized iteratively with the primal-dual method.
		
		\item{\bf{Blocking assumption: observe-think-act paradigm}}
		The environment is assumed to be stationary in DRL while the agents observing the states, computing the learning process, and taking the actions. With limited local computational capability, the processing delay for obtaining the actions in IIoT devices can hardly be negligible. As a result, the environment may have already changed when the device performs the action in a dynamic massive IIoT scenario, and hence the blocking assumption may not hold. One possible solution is to concurrently observe the states and computing the output by formulating continuous-time Bellman equations.

		\item{\bf{Distributed multi-agent DRL}}
		With limited communication resources and a large number of devices, it is very difficult to train a centralized DRL that takes the states and actions of all the devices into account. One promising approach in massive IIoT scenario is to train a multi-agent DRL in a distributed manner, where each agent only observes part of the state of the whole system. However, it is still challenging to find optimal global policy for POMDP due to the following reasons: 1) when all the devices cooperate with each other to maximize a common utility function, the multi-agent DRL may take a long time to converge to a locally optimal policy without QoS guarantee; and 2) when each device tries to maximize its own utility, the system may end up with an equilibrium that is not Pareto optimal (i.e., all the devices experience poor performance). How to improve the convergence performance of multi-agent DRL for POMDP remains a challenging problem, especially for non-stationary environment due to frequent interactions among multiple agents.
	\end{itemize}
	
	{\bf{Integrating DRL with other machine learning:}} To enhance the performance of multi-agent DRL for POMDP, we can combine the DRL with the other machine learning algorithms in above subsections:
	\begin{itemize}
		\item To improve the convergence performance of multi-agent DRL, GNN is a promising technique to enhance the cooperation among agents in the non-stationary environment, where the graph convolution adapts to the dynamic multi-agent environment and capture the interplay between the agents.
		\item To improve the learning performance of DRL in POMDP problems, one promising approach is to utilize the long historical observations. We can integrate the RNN into DRL to extract sufficient features from historical observations to make better decisions.
		\item To ensure the collaboration among all agents, FL framework can be applied in multi-agent DRL. Knowing that the working environments and computation capabilities vary for different devices or BSs, the parameter server in FL should not treat all of them as equal. Specifically, in each FL communication round, the parameter server needs to schedule devices or BSs to upload their local gradients according to their importance. How to design a scheduling policy and define the importance of different devices or BSs remain open problems.
		\item Different from DRL that only estimates the expectation of the value function, a distributional DRL can approximate the distribution of the value function by using GAN \cite{hua2019gan}. The distributional DRL is more suitable for mission-critical IoT, since we are interested in the tail probability of delay rather than the average delay.
		\item Few-shot learning plays a critical role in DRL that needs to adapt to non-stationary environments. As demonstrated in a multi-agent competitive environment in \cite{al2018continuous}, meta-learning is more efficient than reactive baselines in the few-shot regime. Nevertheless, how to guarantee the QoS requirements of critical IIoT with few-shot learning deserves further study.
	\end{itemize}

	\section{Machine Learning Applications}
	Based on the machine learning methods analyzed in Section \ref{IIoT_challenegs}, we present how different machine learning solutions can be applied in individual layer and cross-layer to guarantee the QoS requirements of massive IIoT in this section.
	
	\subsection{Physical Layer}

	Channel prediction and active user detection (AUD) are two important problems in the physical layer: 1) for mobile IoT use cases with predetermined trajectories, it is possible to predict large-scale channel gains for higher resource utilization efficiency and QoS performance. To exploit long-term dependency of historical channel gains, RNN has been utilized in \cite{zhuRNN} for real-time channel prediction; and 2) grant-free transmission can support massive IIoT by enabling data transmission without scheduling, and hence, the BS needs to perform AUD. FNN has been utilized in \cite{KimFCNNgrantfree} for AUD with much lower computational latency than the compressive sensing method. But the time correlation of user activity (i.e., the device transmits over continuous time slots) was not considered, which may be solved by RNN.

	\subsection{MAC Layer}
	Several MAC-layer problems can be formulated as POMDP: 1) to mitigate the serious collision in massive IoT access, DRL can be applied to optimize access control schemes. In \cite{jiang2020traffic, jiang2020decoupled}, DRL dynamically adapts the access control factors based on the traffic prediction via RNN. With multi-agent configuring the parameters of different schemes, the GNN has the potential to further enhance their cooperation; and 2) scheduling algorithms are important components to provide the guaranteed IIoT QoS requirements. The authors of \cite{gu2020knowledge} proposed a knowledge-assisted DRL algorithm to optimize the scheduling policy, where the expert knowledge is exploited to improve the training efficiency of DRL. To optimize the scheduler in more general scenarios with dynamic user requests and diverse QoS requirements, the combination of constraint DRL with GNN is a potential solution.

	\subsection{Network Layer}
	Machine learning methods have been applied to deal with network-layer problems in recent studies, including network slicing and routing: 1) to meet the diverse QoS requirements, learning-based network slicing is a promising approach. The authors of \cite{hua2019gan} developed a GAN-powered deep distributional RL for resource management in network slicing, where GAN is used to approximate the action-value distribution. However, the scheduling policy was assumed to be Round Robin and the achievable rate was characterized by Shannon's capacity for simplicity; and 2) in long-distance communication scenarios, like UAV communications, routing delay is one of the major components of E2E delay. \cite{rusek2020routenet} applied GNN to evaluate the latency, jitter, and packet losses of a routing scheme, and indicated that by combining GNN with DRL, it is possible to minimize the E2E latency, jitter, or packet losses.

	\subsection{Cross-layer Design}

	To improve the E2E performance, we should optimize the system in a cross-layer manner. Different from existing approaches that divide the system into multiple layers, cross-layer models are more complicated and the optimization problems are non-convex in general. To overcome this difficulty, we can integrate theoretical knowledge of different layers (models, analysis tools, and optimization frameworks) into deep learning algorithms \cite{she2020tutorial}. Specifically, DNNs are first pre-trained offline in a simulation platform built upon theoretical knowledge and then fine-tuned in real-world systems to handle the mismatch between simulation and real-world systems.

	\section{Case study}
	
	In massive IIoT, the BS, with limited preambles $F = 54$, is required to optimize the massive access to mitigate the collisions among devices. Therefore, we validate the effectiveness of machine learning algorithms in optimizing massive access problems under two typical traffic types: one shot static traffic scenario and continuous traffic scenario.

	\subsection{One Shot Static Traffic Scenario}
	IIoT devices transmit their latest measurement to the BS periodically, and the BS determines each device to transmit or not in each time slot. To maximize the number of successful transmissions in each time slot, we model this problem as unsupervised learning, where the perfect CSIs are assumed to be known at the BS. With a total number of $ \rm{N_{d}=80} $ devices, we take each device as a node and the CSI as an edge between the device and the BS, and characterize the system by a $ \rm{N_{d}}\times\rm{N_{d}}  $ diagonal CSI matrix \cite{Mark2020Optimal}. For GNN and CNN, we have $L = 10$ layers and 40 parameters in total, and dropout mechanism is applied to overcome the over-smoothing problem. For the one layer RNN with 128 neurons, the input is each row of the matrix, and the total number of parameters is around $10^5$. For the FNN with two hidden layers of size 64 and 32, the total number of the parameters is around $4\times 10^5$.
	\begin{figure}[!htb]
		\centerline{\includegraphics[scale=0.4]{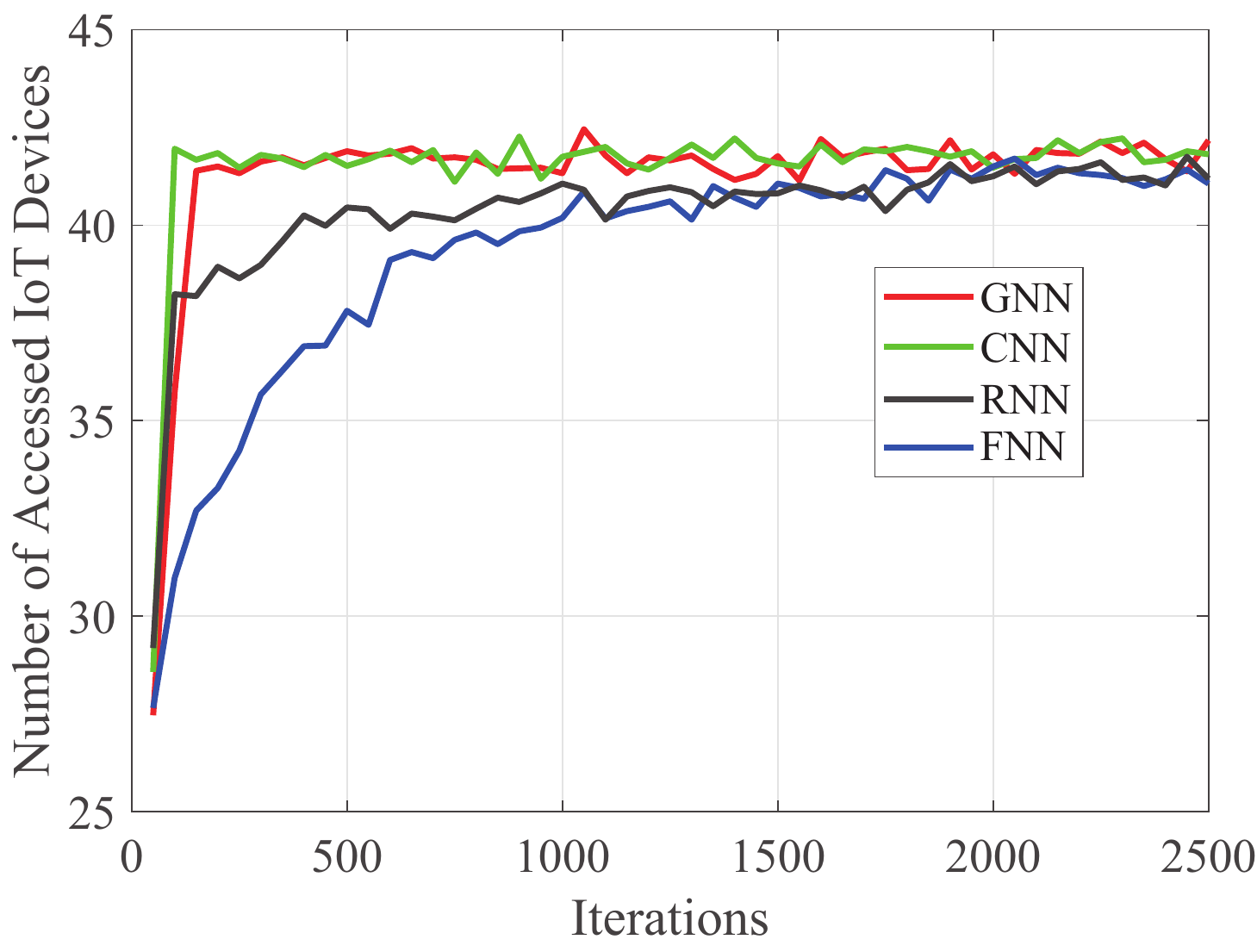}}
		\caption{Average number of successfully accessed IoT devices with static traffic.}
		\label{fig:Success number period}
	\end{figure}
	
	Fig.~\ref{fig:Success number period} compares the performance of GNN, CNN, FNN, and RNN by evaluating the average number of successful devices in random access. Our results shown that all of them achieve similar performance after convergence, and the convergence speed follows GNN $\approx$ CNN $>$ RNN $>$ FNN. This is because the number of parameters of the FNN and RNN is much larger than the number of parameters of the GNN and CNN in massive IIoT problems.
	
	\subsection{Continuous Traffic Scenario} 
	IIoT devices transmit sporadic traffic to the BS, and the BS adopts Access Class Barring (ACB) scheme to optimize the uplink transmission. Since the number of active devices is determined by the newly arrived packets and previously failed packets, we model this problem as a Markov decision problem to maximize the number of successful transmissions over the long term. We set the number of devices and the maximum number of retransmissions as $ \rm{N_{d}=500} $ and $ \mathrm{N_{r}=10}$, respectively. The reward in each slot is defined as the number of successful transmissions. DQN and DDPG are applied to optimize the ACB factor ($\theta \in (0, 1])$) in the discrete (i.e., with the pace of 0.05) and continuous action space \cite{jiang2020decoupled}, respectively.

	\begin{figure}[!htb]
		\centerline{\includegraphics[scale=0.4]{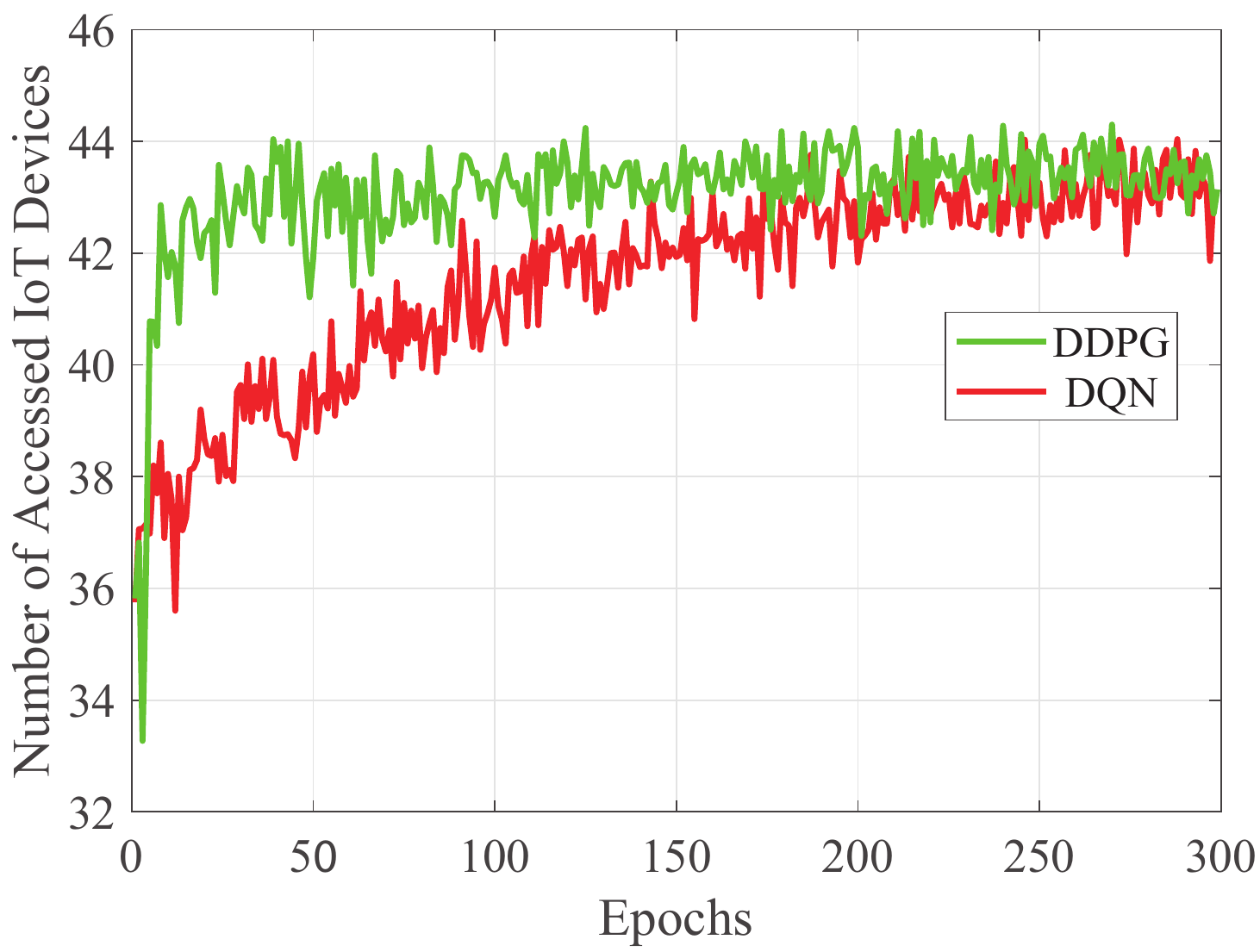}}
		\caption{Average number of successfully accessed IoT devices with continuous traffic.}
		\label{fig:Success number sporadic}
	\end{figure}
	Fig.~\ref{fig:Success number sporadic} compares the average number of success devices in the random access of DQN and DDPG. We can observe that the DDPG slightly outperforms the DQN. This is due to the  of DDPG utilizes continuous control mechanism, which provides the infinite action space for the ACB configuration, whereas the action of DQN is limited to 20 discrete values. 
	
	\section{Conclusion}
	In this article, we summarized the QoS requirements of massive IIoT use cases in a Beyond 5G enabled factory. We then focused on investigating machine learning solutions to deal with massive IIoT problems with specific characteristic, along with its limitations and potential research directions. Based on the machine learning solutions, We further reviewed how to apply them to guarantee the QoS requirements of massive IIoT in individual layer, and cross-layer design, respectively. Importantly, we utilized different DNN structures and DRL in massive access optimization problem to validate the effectiveness of the machine learning solutions. This work serves to inspire research to customize machine learning solutions that consider characteristics of massive IIoT with stringent QoS requirements.
	
	%



	

	\ifCLASSOPTIONcaptionsoff
	\newpage
	\fi


	
	\bibliographystyle{IEEEtran}
	\bibliography{arxiv_version}
	\vskip -2\baselineskip plus -1fil
	
	\begin{IEEEbiographynophoto}{Hui Zhou}
		is currently a Ph.D. student in the Center for Telecommunications  Research (CTR), King's College London.
	\end{IEEEbiographynophoto}
	
	\vskip -2\baselineskip plus -1fil
	\begin{IEEEbiographynophoto}{Changyang She}
		is currently a postdoctoral research associate at the University of
		Sydney. His research interests lie in the areas of ultra-reliable and low-latency
		communications, tactile internet, mobile edge computing, internet-of-things,
		deep learning in 5G and Beyond.
	\end{IEEEbiographynophoto}

	\vskip -2\baselineskip plus -1fil
	
	\begin{IEEEbiographynophoto}{Yansha Deng} 
		is currently a  Lecturer  (Assistant  Professor) in the  CTR, King’s College London. Her research interests include machine learning for 5G/B5G wireless networks and molecular communications. 
	\end{IEEEbiographynophoto}
	
	\vskip -2\baselineskip plus -1fil
	
	\begin{IEEEbiographynophoto}{Mischa Dohler} 
		is a full professor in wireless communications with King’s College London, driving cross-disciplinary research and innovation in technology, sciences, and arts. He is
		a Fellow of the Royal Academy of Engineering, the Royal Society of Arts (RSA), and the Institution of Engineering and Technology (IET), and a Distinguished Member of Harvard Square Leaders Excellence.
	\end{IEEEbiographynophoto}

	\vskip -2\baselineskip plus -1fil
	\begin{IEEEbiographynophoto}{Arumugam Nallanathan}
		is Professor of Wireless Communications and Head of the Communication Systems Research (CSR) group in the School of Electronic Engineering and Computer Science at Queen Mary University of London. His research interests include Beyond 5G Wireless Networks, Internet of Things, and Molecular Communications.
	\end{IEEEbiographynophoto}

	%
	%
	
	
	
\end{document}